\documentclass[letterpaper, 10 pt, conference]{ieeeconf}  

\IEEEoverridecommandlockouts                              

\usepackage{amsmath} 
\usepackage{amssymb}  
\usepackage[linesnumbered,ruled]{algorithm2e}
\usepackage{gensymb}
\usepackage{graphicx}
\usepackage{epstopdf}
\usepackage{subcaption}
\usepackage[colorlinks]{hyperref}
\usepackage{multirow}
\usepackage[T1]{fontenc}
\usepackage[super]{nth}
\usepackage{fix-cm}
\usepackage{floatrow}
\usepackage[]{caption}

\makeatletter
\let\NAT@parse\undefined
\makeatother

\usepackage[square,comma,numbers,sort&compress]{natbib} 

\title{\LARGE \bf
Deep Learning for Spacecraft Pose Estimation from Photorealistic Rendering
}

\author{Pedro F. Proen\c{c}a$^{1}$ and Yang Gao$^{1}$
\thanks{$^{1}$The authors are with the Surrey Space Centre, Faculty of Engineering and Physical Sciences, University of Surrey, GU2 7XH Guildford,
	U.K. {\tt\small \{p.f.proenca, yang.gao\}@surrey.ac.uk}}%
}

\begin{document}

\maketitle
\thispagestyle{empty}
\pagestyle{empty}

\begin{abstract}
On-orbit proximity operations in space rendezvous, docking and debris removal require precise and robust 6D pose estimation under a wide range of lighting conditions and against highly textured background, i.e., the Earth.

This paper investigates leveraging deep learning and photorealistic rendering for monocular pose estimation of known uncooperative spacecrafts. 
We first present a simulator built on Unreal Engine 4, named URSO, to generate labeled images of spacecrafts orbiting the Earth, which can be used to train and evaluate neural networks. 

Secondly, we propose a deep learning framework for pose estimation based on orientation soft classification, which allows modelling orientation ambiguity as a mixture of Gaussians. This framework was evaluated both on URSO datasets and the ESA pose estimation challenge. In this competition, our best model achieved \nth{3} place on the synthetic test set and \nth{2} place on the real test set. Moreover, our results show the impact of several architectural and training aspects, and we demonstrate qualitatively how models learned on URSO datasets can perform on real images from space.

\end{abstract}



\section{Introduction}

Spacecraft position and attitude estimation is essential to on-orbit operations \cite{nanjangud2018robotics}, e.g., formation flying, rendezvous, docking, servicing and space debris removal \cite{taylor2018remove}. These rely on precise and robust estimation of the relative pose and trajectory of object targets in close-proximity under harsh lighting conditions and against highly textured background (i.e. Earth). As surveyed in \cite{opromolla2017review}, according to the specific operation scenario, the targets may be either: (i) cooperative if they use a dedicated radio-link, fiducial markers or retro-reflectors to aid pose determination or (ii) non-cooperative with either unknown or known geometry. Recently, the latter has been gaining more interest by both the research community and space agencies due mainly to the accumulation of inactive satellites and space debris in low Earth orbit \cite{forshaw2016removedebris} but also military space operations. For instance, ESA opened a competition \cite{ESAChallenge}, this year, to estimate the pose of a known spacecraft from a single image using supervised learning. This papers addresses this problem.
\par
\begin{figure}[t]
\centering
	\begin{tabular}{@{}c@{ }c@{}}
		\includegraphics[width=40.5mm,height=30.5mm]{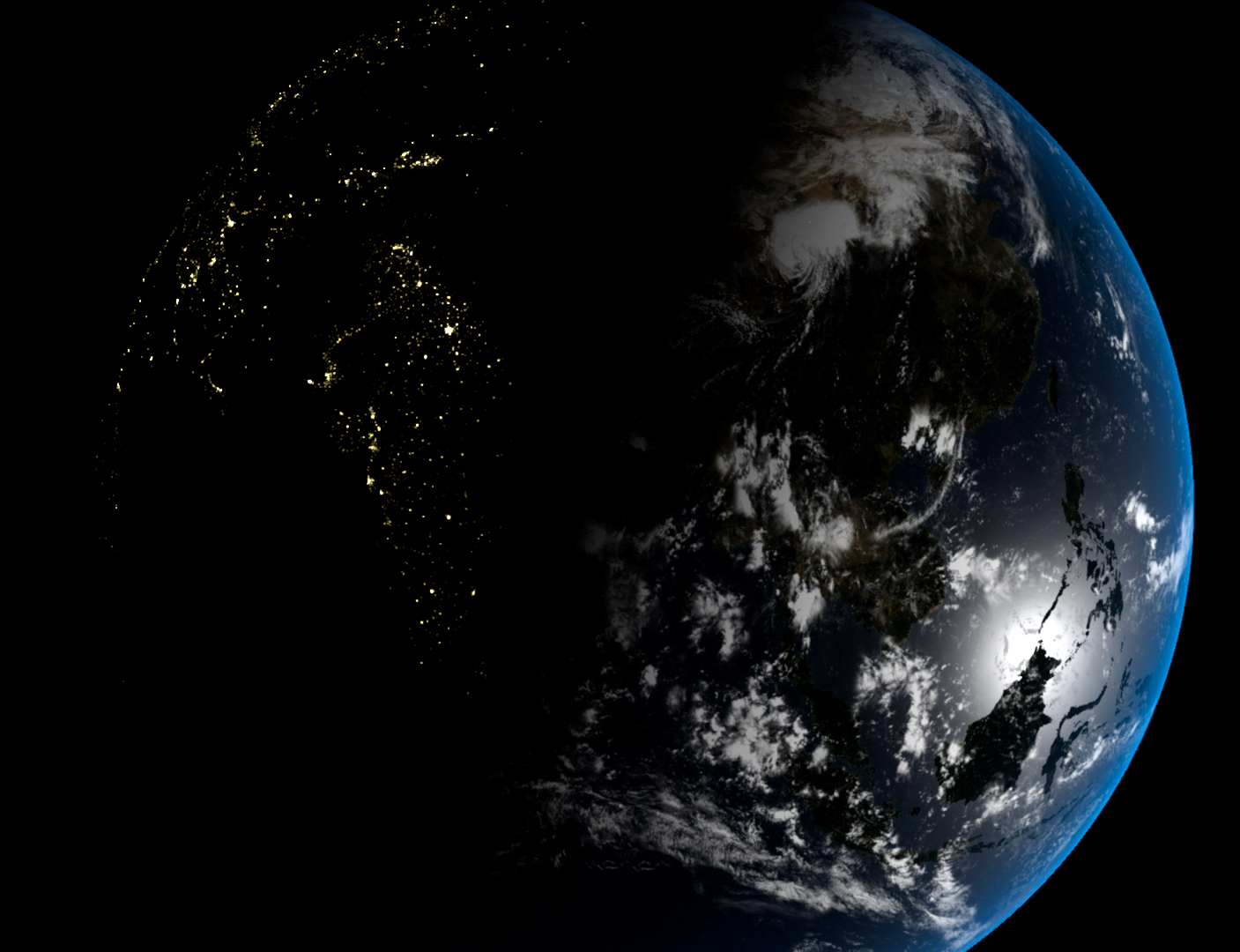} &
		\includegraphics[scale=0.09]{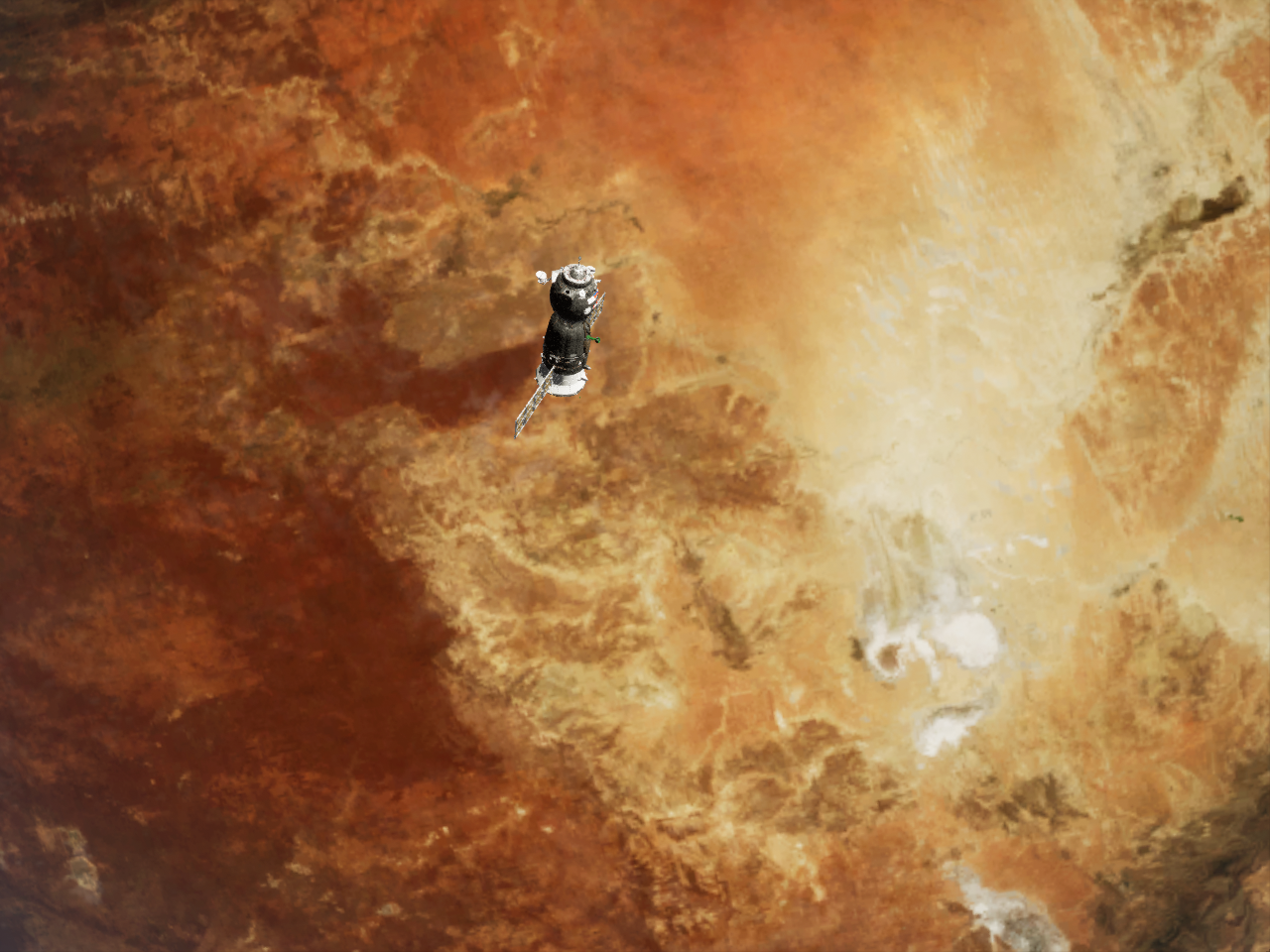}\\
		\includegraphics[scale=0.09]{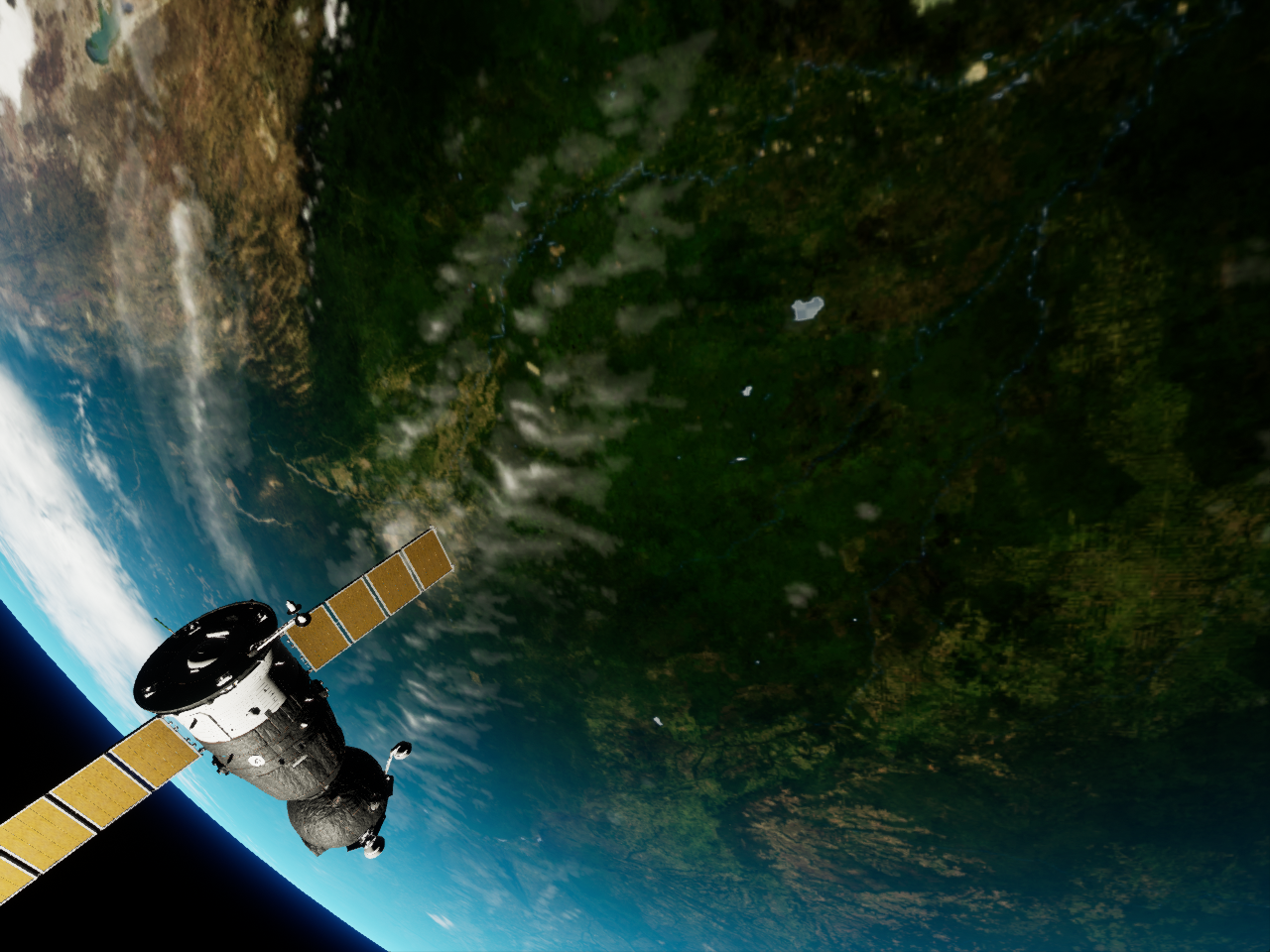} &
		\includegraphics[scale=0.09]{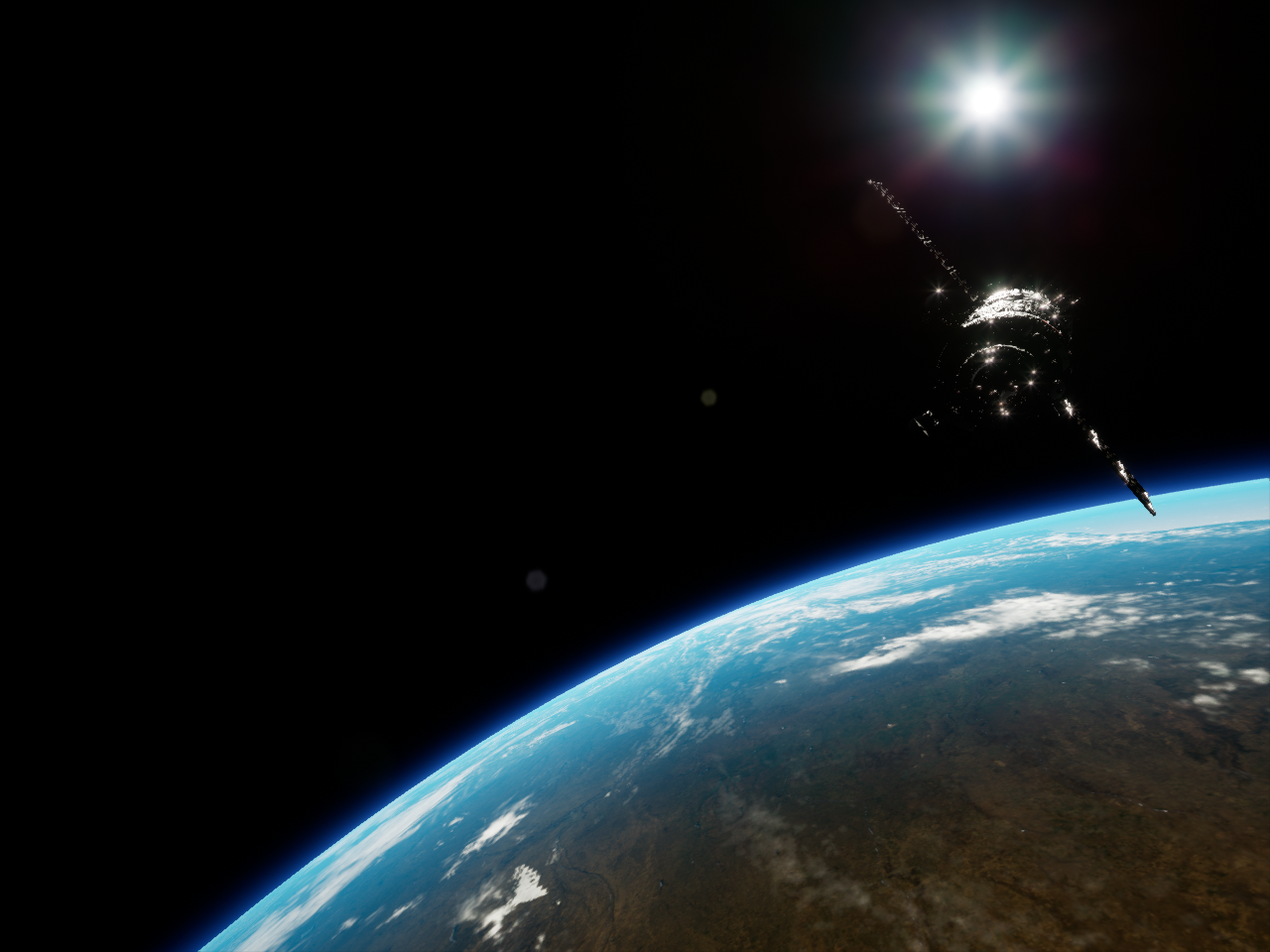}\\
	\end{tabular} 
	\caption{ Example of frames synthesized by URSO of a soyuz model. For videos and the datasets used in this work, refer to: {\small \url{https://pedropro.github.io/project/urso/}}}
	\vspace*{-1mm} 
	\label{fig:teaser}
\end{figure}

The main limitation of deep learning (DL) is that it needs a lot of data, which is especially costly in space. Therefore, as our first contribution, we propose a visual simulator built on Unreal Engine 4, named URSO, which allows obtaining photorealistic images and depth masks of commonly used spacecrafts orbiting the Earth, as seen in Fig. \ref{fig:teaser}.
Secondly, we carried out an extensive experimental study of a DL-based pose estimation framework on datasets obtained from URSO, where we investigate the performance impact of several aspects of the architecture and training configuration. Among our findings, we conclude that data augmentation with random camera orientation perturbations is quite effective to combat overfitting and we present a probabilistic orientation estimation via soft classification that performs significantly better than direct orientation regression and it can further model uncertainty due to orientation ambiguity as a Gaussian mixture. Moreover, our best solution achieved \nth{3} place on the synthetic dataset and \nth{2} place on the real dataset of ESA pose estimation challenge \cite{ESAChallenge}. We also demonstrate qualitatively how models trained on URSO data can generalize to real images from space through our augmentation pipeline. 

\section{Related Work}
Previous monocular solutions \cite{naasz2010flight,kelsey2006vision,liu2014relative,petit2013robust,petit20153d, capuano2019robust} to spacecraft tracking and pose estimation rely on model-based approaches (e.g. \cite{DrummondCipolla}) that align a wireframe model of the object to an edge image (typically given by a Canny detector) of the real object based on heuristics. However objects are more than just a collection of edges and geometric primitives. Convolutional Neural Networks (CNNs) can learn more complex and meaningful features to the task at hand while ignoring background features (e.g. clouds) based on context.

Despite the maturity of DL in many computer vision tasks. Only recently \cite{kendall2015posenet,xiang2018posecnn,kehl2017ssd,do2018deep,hu2019segpose,tekin2018real,rad2017bb8} has DL become common in pose estimation problems. 
Kendall \textit{et al.} \cite{kendall2015posenet} first proposed adapting and training GoogLeNet on Structure-from-Motion models for camera relocalization. Their network was trained to regress a quaternion by minimizing the $L_2$ loss between quaternions. Moreover, they extended their method to model uncertainty by using Monte Carlo sampling with dropout. \cite{kendall2016modelling}. Kehl \textit{et al.} \cite{kehl2017ssd} proposed a DL solution for detection and pose estimation of multiple objects based on hard viewpoint classification, where ambiguous views are manually removed a-priori. On the other hand, Xiang \textit{et al.} \cite{xiang2018posecnn} proposed a model based on a segmentation network for handling multiple objects. While object locations are estimated using Hough voting on the network image output, their orientations are estimated through quaternion regression following ROI pooling. To account for object symmetries, a loss function based on ICP is used, but this is prone to local minima and it requires a depth map. To handle more than one object instance per class, Thanh-Toan \textit{et al.} \cite{do2018deep} extended Mask-RCNN to pose estimation by simply adding a head branch, which regresses orientation as the angle-axis vector. Although, this minimal parameterization avoids the quaternion normalization, they still employed an $L_2$ loss function. Mahendran \textit{et al.} \cite{mahendran20173d} also regress the angle-axis vector, but they minimize directly the geodesic loss. Su \textit{et al.}\cite{su2015render} performed fine grained hard viewpoint classification. Hara \textit{et al.} \cite{hara2017designing} compared regressing the azimuth using either the $L_2$ loss or the angular difference loss versus hard classification with mean-shift algorithm to retrieve a continuous value. DL has also been successfully applied to visual odometry \cite{wang2017deepvo,Zhou_2018_ECCV}. While Wang \textit{et al.} \cite{wang2017deepvo} simply regresses Euler angles, Zhou \textit{et al.} \cite{Zhou_2018_ECCV} regress simultaneously multiple (i.e. 64) pose hypothesis with angle-axis representation and then average them since pose updates in visual odometry are usually small. There is a large body of work on pose estimation from RGB-D images, which was recently comprehensively evaluated in \cite{hodan2018bop}, where typically ICP is used for pose refinement. In their benchmark, \cite{hodan2018bop} concluded that learning-based solutions are still not on par with point-cloud-based methods \cite{hinterstoisser2016going} in terms of precision. But more recently, \cite{hu2019segpose,tekin2018real,rad2017bb8} have advanced state-of-the-art by refraining from estimating directly pose and instead use CNNs to regress the 2D projections of predefined 3D keypoints and finally estimate pose using robust P$n$P solutions, e.g., embedded in RANSAC. Approaches such as \cite{hu2019segpose} however need further work to handle very small or far-away objects, as they rely on coarse segmentation grids.

\par
Sharma \textit{et al.} \cite{sharma2018pose} were the first to propose using CNNs for spacecraft pose estimation based on hard viewpoint classification, but later they \cite{sharma2019pose} proposed doing position estimation based on bounding box detection and orientation estimation based on soft classification. Although, the approach to position fails when part of object is outside the field of view, the orientation estimation has its merits. Two head branches are used for orientation estimation: one does hard classification, given a set of pre-defined quaternions, to find the $N$ closest quaternions, then a second branch estimates the weights for these $N$ quaternions, and the final orientation is given by the weighted average quaternion. Our method for orientation estimation is similar to this approach, however, our framework does not require two orientation branches, provides intuitive regularization parameters and can handle multiple hypothesis due to perceptual aliasing. \par 
The same work \cite{sharma2019pose} introduced the dataset used in the ESA challenge, which is just made of montages of real images with basic OpenGL renderings of a satellite. On the other hand, two image simulation tools have so far been specifically developed to support vision-based navigation in space scenes (e.g. Martian surface, asteroid landing): the early PANGU \cite{parkes2004planet} used by ESA and the more comprehensive Airbus internal simulator: SurRender \cite{brochard2018scientific}, which supports ray tracing and very large datasets. Nevertheless, state-of-the-art game engines (e.g. UE4), widely used in autonomous driving \cite{Dosovitskiy17} and robotics \cite{shah2018airsim,martinez2018unrealrox} offer far more resources to develop complex and photorealistic environments, but these have been criticized in \cite{brochard2018scientific} for being designed for human vision and lacking the photometric accuracy of actual sensors. We point out that recent efforts have been made in the source-available UE4 to implement physically-based shading models and cameras.

\begin{figure}[t]
	\centering
	\includegraphics[scale=0.49]{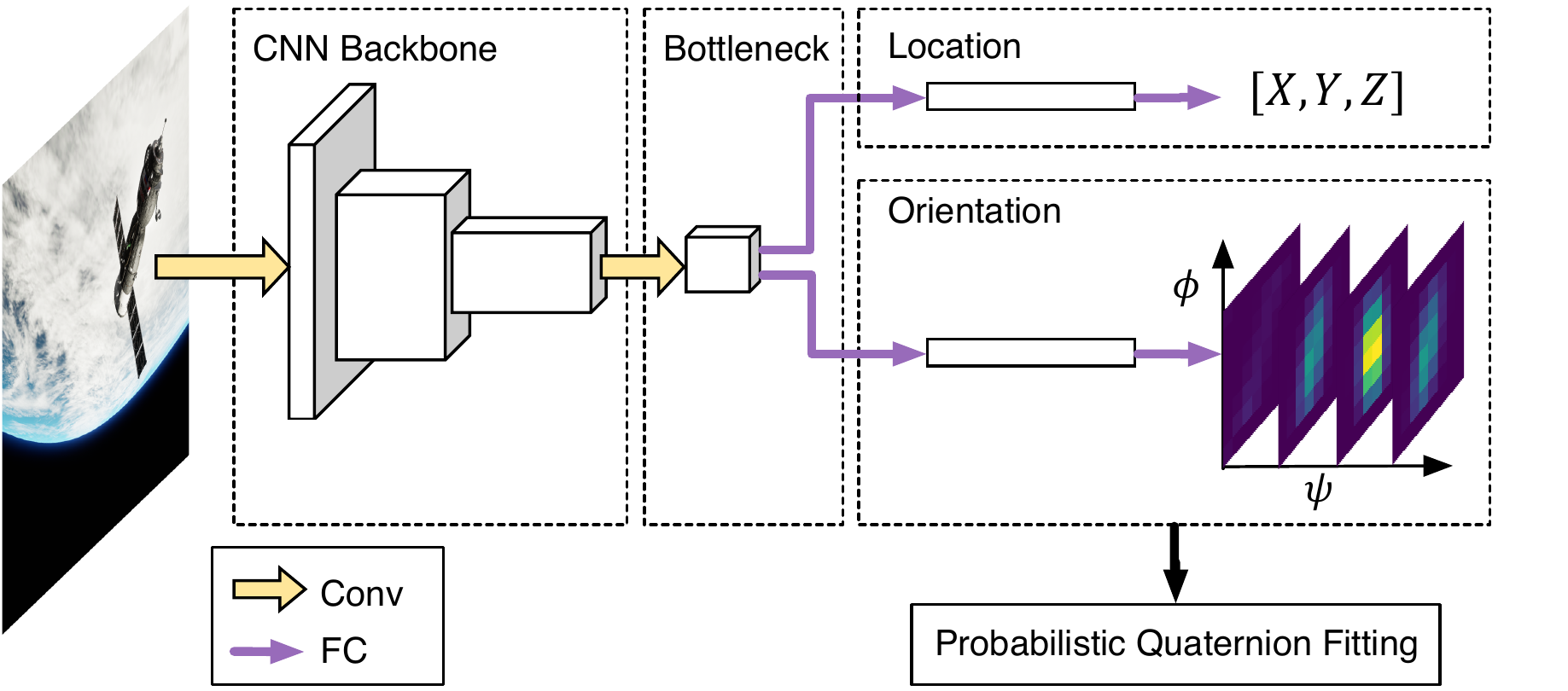}
	\caption{Simplified overview of the network architecture proposed in this work.}
	\label{fig:pipeline}
\end{figure}

\section{Pose Estimation Framework}
\label{sec:method}

Our network architecture, depicted in Fig. \ref{fig:pipeline} is aimed at simplicity rather than efficiency to perform a first ablation study. We adopted the ResNet architectures with pre-trained weights as the network backbone, due to its low number of pooling layers, and good accuracy-complexity trade-off \cite{canziani2016analysis}. The last fully-connected layer and the global average pooling layer of the original network were removed to keep spatial feature resolution, leaving effectively only one pooling layer at the second layer. The global pooling layer was replaced by one extra 3$\times$3 convolution with stride of 2 (bottleneck layer) to compress the CNN features since our task branches are fully-connected to the input tensor. For lower space complexity, one could use instead a Region Proposal Network as in \cite{xiang2018posecnn,do2018deep,he2017mask}, but this complicates our end-to-end pose estimation. As a drawback, our network does not handle multiple objects \textit{per se}.
\par

Our 3D location estimation is a simple regression branch with two fully-connected layers, but instead of minimizing the absolute Euclidean distance, we minimize the relative error, corresponding to the first term of our total loss function:

\begin{equation}
\label{eq:loc_total}
L_{\textrm{total}} = \beta_1 \sum_{i}^{m}\frac{\|t^{(i)}-t^{(i)}_{gt}\|_2}{\|t^{(i)}_{gt}\|_2} + \beta_2 L_{\textrm{ori}}
\end{equation}
where $t^{(i)}$ and $t^{(i)}_{gt}$ are respectively the estimated and ground-truth translation vector. The solely advantage of minimizing the relative error, is that the fine-tuned loss weights $\{\beta_1,\beta_2\}$ in our experiments generalize better to other datasets, as this loss does not depend on the translation scale. To avoid having to fine-tune loss weights, we have also experimented instead in Section \ref{sec:experiments} regressing three virtual 3D keypoints and then estimate pose using a closed-form solution \cite{arun1987least}.

\subsection{Direct Orientation Regression}
\label{sec:ori_reg}

While several works \cite{kendall2015posenet,do2018deep,wang2017deepvo} have used $L_2$ or $L_1$ loss to regress orientation. This does not represent correctly the actual angular distance for any orientation representation. Quaternions, for example, are non-injective. While one can map quaternions to lie only on one hemisphere as in \cite{kendall2017geometric}, $L_2$ distances to quaternions near the equator will still not express the geodesic distance. Therefore we have experimented minimizing directly either:
$L_{\alpha}=\arccos(|{q^{(i)}}^\top q_{gt}^{(i)}|)$
or:
$L_{\cos{\alpha}}=1-|{q^{(i)}}^\top q_{gt}^{(i)}|$ to regress a unit quaternion $q^{(i)}$, subject to a normalization layer. One possible issue with the first expression is that the derivative of $\cos^{-1}(x)$ is infinite at $x=1$, but this can be easily solved by scaling down $x$.

\subsection{Probabilistic Orientation Soft Classification}

Alternatively, we propose to do continuous orientation estimation via classification with soft assignment coding \cite{liu2011defense}. The key idea is to encode each label ($q_{gt}$) as a Gaussian random variable in an orientation discrete output space (represented in Fig. \ref{fig:pipeline}), so that the network learns to output probability mass functions. To this end, a 3D histogram is used as the network output, where each bin maps to a combination of discrete Euler angles specified by the quantization step. Special care is taken to avoid redundant bins in the \textit{Gimbal lock} and borders. Let $Q=\{b_1,..,b_N\}$ be the quaternions corresponding to the histogram bins, then, during training, each bin is encoded with the soft assignment function:
\begin{equation}
\label{eq:soft_encoding}
f(b_i,q_{gt}) = \frac{K(b_i,q_{gt})}{\sum_{j}^{N}K(b_j,q_{gt})} 
\end{equation}
where the kernel function $K(x,y)$ uses the normalized angular difference between two quaternions:
\begin{equation}
\label{eq:kernel_fx}
K(x,y)= e^{-\frac{\big(\frac{2\cos^{-1}(|x^\top y |)}{\pi}\big)^2}{2 \sigma^2}}
\quad \textrm{and} \quad
\sigma^2 = \frac{\big(\frac{\Delta}{M}\big)^2}{12}
\end{equation}
 and the variance $ \sigma^2$ is given by the quantization error approximation, where $\Delta/M$ represents the quantization step, $\Delta$ is the smoothing factor that controls the Gaussian width and $M$ is the number of bins per dimension (i.e. Euler angle).

At test time, given the bin activations $\{a_1,..,a_N\}$ and the respective quaternions, in one hemisphere, we can fit a quaternion by minimizing the weighted least squares:
\begin{equation}
\label{eq:quat_avg}
\hat{q} = \operatorname*{argmin}_q  \sum_{i}^{N} w_i (1-{b_i}^\top q)^2
\end{equation}
where $a_i$ is assigned to $w_i$ and the optimal solution is given by the right null space of the matrix $\sum_{i}^{N} w_i ({b_i}{b_i}^\top)$ \cite{markley2007averaging}. This solution was also employed in \cite{sharma2019pose}.
\par

\subsection{Multimodal Orientation Estimation}
\label{multimodal_EM}

When there are ambiguous views in the training-set, this results in one-to-many mappings, therefore the optimal network that minimizes the cross entropy losses, given the soft assignments in (\ref{eq:soft_encoding}), will output a multimodal distribution.
To extract multiple orientation hypothesis from such network's output, we propose an Expectation-Maximization (EM) framework to fit a Gaussian Mixture model $\Theta=\{\theta_1,...,\theta_K\}$ with means $\{q_1,...,q_K\}$.
As the E step, for every model $\theta_j$ and bin we compute the membership:
\begin{equation}
\label{eq:E_step}
p(\theta_j|b_i) = \frac{p(b_i|\theta_j) p(\theta_j)}{\sum_{k}^{K}p(b_i|\theta_k) p(\theta_k)}
\end{equation}
where $p(b_i|\theta_j)= K(b_i,q_j)$ with $\sigma_j$ initialized as in (\ref{eq:kernel_fx}) and the priors $p(\theta_j)$ as equiprobable. These are then updated in the M step:
\begin{equation}
\label{eq:prior}
p(\theta_j) = \sum_{i}^{N} a_i p(\theta_j|b_i) \textrm{ {\small and} } 
\sigma_{j} = \sum_{i}^{N} w_{ji} \Big(\frac{2\cos^{-1}(|b_i^\top q_j |)}{\pi}\Big)^2
\end{equation}
where $q_j$ is firstly obtained by solving (\ref{eq:quat_avg}) with the weights:
$
w_{ji} = \frac{a_i p(\theta_j|b_i)} {p(\theta_j)}
$
. The model means are initialized as the $K$ bins with strongest activations after non-maximum suppression.
To find the optimal number of models, we increase $K$ until the log-likelihood stops increasing by more than a threshold.

\section{URSO: Unreal Rendered Spacecraft On Orbit}

Our simulator leverages Unreal Engine 4 (UE4) features to render realistic images, e.g., physically based materials, bloom and lens flare. Lighting in our environment is simply made of a directional light and spotlight to simulate respectively sunlight and Earth albedo. Ambient lighting was disabled and to simulate the sun we used a body of emissive material with UE4 bloom scatter convolution. Earth was modelled as a high polygonal sphere textured with $21600 \times 10800$ Earth and cloud images from the Blue Marble Next Generation collection \cite{BlueMarble}. This is further masked to obtain specular reflections from the ocean surface. Additionally a third party asset is used to model the atmospheric scattering. Our scene includes a Soyuz and Dragon spacecraft models with geometry imported from 3D model repositories \cite{Turbosquid}. \par
To generate datasets, we sample randomly $5000$ viewpoints around the day side of the Earth from low Earth orbit altitude. The Earth rotation, camera orientation and target object pose are all randomized. Specifically, the target object is placed randomly within the camera viewing frustum and an operating range between [10,40] m. Our interface uses UnrealCV plugin \cite{qiu2017unrealcv}, which allows obtaining an RGB image and depth map for each viewpoint. Images were rendered at a resolution of 1080$\times$960 pixels by a virtual camera with a 90$^\circ$ horizontal FOV and auto-exposure.

\section{Data Augmentation and Sim-to-Real Transfer}

\begin{figure}[tb]
	\centering
	\begin{tabular}{@{}c@{ }c@{ }c@{}}
		\includegraphics[scale=0.18]{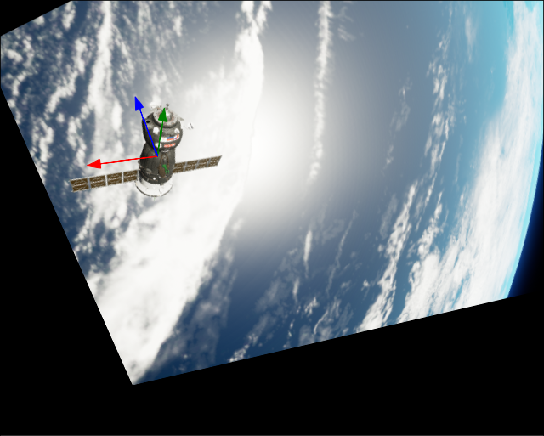}&
		\includegraphics[scale=0.44]{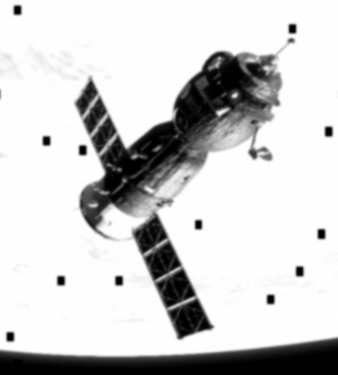}&
		\includegraphics[scale=0.44]{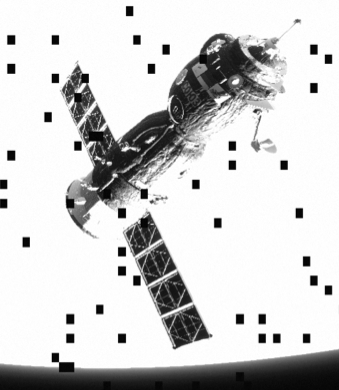}\\[-5pt]
		\scriptsize{(a)} & \scriptsize{(b)} & \scriptsize{(c)} \\
	\end{tabular}
	\begin{tabular}{@{}c@{ }c@{}}
		\includegraphics[scale=0.205]{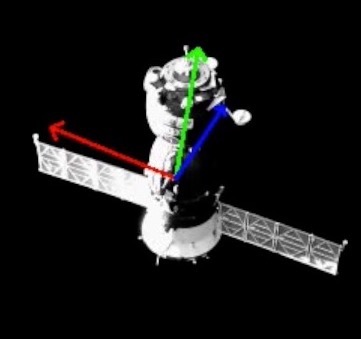} &
		\includegraphics[scale=0.195]{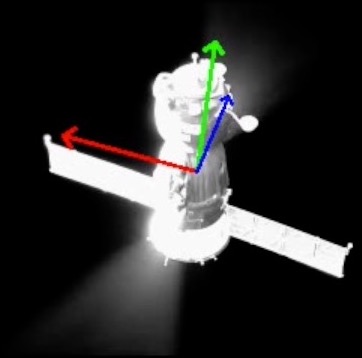} \\
		\scriptsize{(d)} & \scriptsize{(e)}  \\[-5pt]
	\end{tabular}
	\caption{Image augmentation and sim-to-real examples. (a) Image warped due to camera orientation perturbation, (b) and (c) Images after our sim-to-real post-processing. (d) and (e) show real images (5 seconds apart) of a soyuz with overlayed estimated pose after training with data augmentation. Notice the thrusters in action on (e).}
	\label{fig:aug}
\end{figure}

Typical image transformations (e.g. cropping, flipping) have to be considered carefully as these may change the object nature and camera intrinsic parameters, which, in our case, is embedded in the network. One can do random in-plane rotation, since there is no concept of up and down in space, but the object may get out of bounds due to the aspect ratio, therefore this was only done for the ESA \& Stanford dataset, where the satellite is always nearly centered. Additionally, we can cause small random perturbations to the camera orientation by warping the images as shown in Fig. \ref{fig:aug}. We do this during training and accordingly update the pose labels by repeating the encoding in (\ref{eq:soft_encoding}). To generalize the learned models to real data, we convert the images to grayscale, change the image exposure and contrast, add AWG noise, blur the images and drop out patches as shown in Fig. \ref{fig:aug}. The motivation to use the latter is that it can help disentangling features from our mock-up that do not match the real object and it can improve robustness to occlusions and shadows.

\section{Experiments}
\label{sec:experiments}

We conducted experiments on datasets captured using URSO and the ESA \& Stanford's benchmark dataset \cite{ESAChallenge}, named SPEED. The latter contains both synthetic and real images with 1920$\times$1200 px, generated in \cite{sharma2019pose}, of a mock-up model of one satellite used in a flight mission, named PRISMA \cite{d2012spaceborne}. The testing set contains 300 real images and 2998 synthetic images, whereas the training-set contains 12000 synthetic images and only 5 real images. All images are in grayscale. The labels of the testing set are not provided, instead the methods are evaluated by the submission server based on a subset of the testing-set.
As for URSO, we collected one dataset for the dragon spacecraft and two datasets for the soyuz model with different operating ranges: \textit{soyuz\_easy} with [10-20] m and \textit{soyuz\_hard} with [10-40] m. Low ambient light was also exceptionally enabled on \textit{soyuz\_easy}. We have noticed that training on \textit{soyuz\_easy} converges faster, therefore our first experiments in this section use this dataset. All three datasets contain 5000 images, of which 10\% were held out for testing and another 10\% for validation. Performance is reported as the mean absolute location error, the mean angular error and also the metric used by the ESA challenge server, referred to as \textit{ESA Error}, which is the sum of the mean relative location error, as in (\ref{eq:loc_total}), and the mean angular error.

\subsection{Implementation and Training Details}
Networks were trained on one NVIDIA GTX 2080 Ti, using SGD with a moment of 0.9, a weight decay regularization of 0.0001 and a batch size of 4 images. Training starts with weights from the backbone of Mask R-CNN trained on COCO dataset, since we use high image resolutions. The learning rate ($lr$) was scheduled using step decay depending on the model convergence, which we have found to depend highly on the orientation estimation method, number of orientation bins, augmentation pipeline and the dataset. By default, unless explicitly stated, we used: ResNet-50 with a bottleneck width of 32 filters, orientation soft classification with 16 bins per Euler angle, camera rotation perturbations with maximum magnitude of 10$^\circ$  to augment the dataset and images were resized to half their original size. Training a model with this default configuration on \textit{soyuz\_easy} converges after 30 epochs with $lr=0.001$  plus 5 epochs with $lr=0.0001$, whereas orientation regression takes approximately half the number of iterations.

\subsection{Results}
\label{sec:results}

First, results from fine-tuning the parameters of our probabilistic orientation estimation based on soft classification are shown in Table \ref{tab:ori_tuning} for \textit{soyuz\_easy}.

\begin{figure}[b]
	\vspace{-20pt}
	\centering
	\includegraphics[scale=0.55]{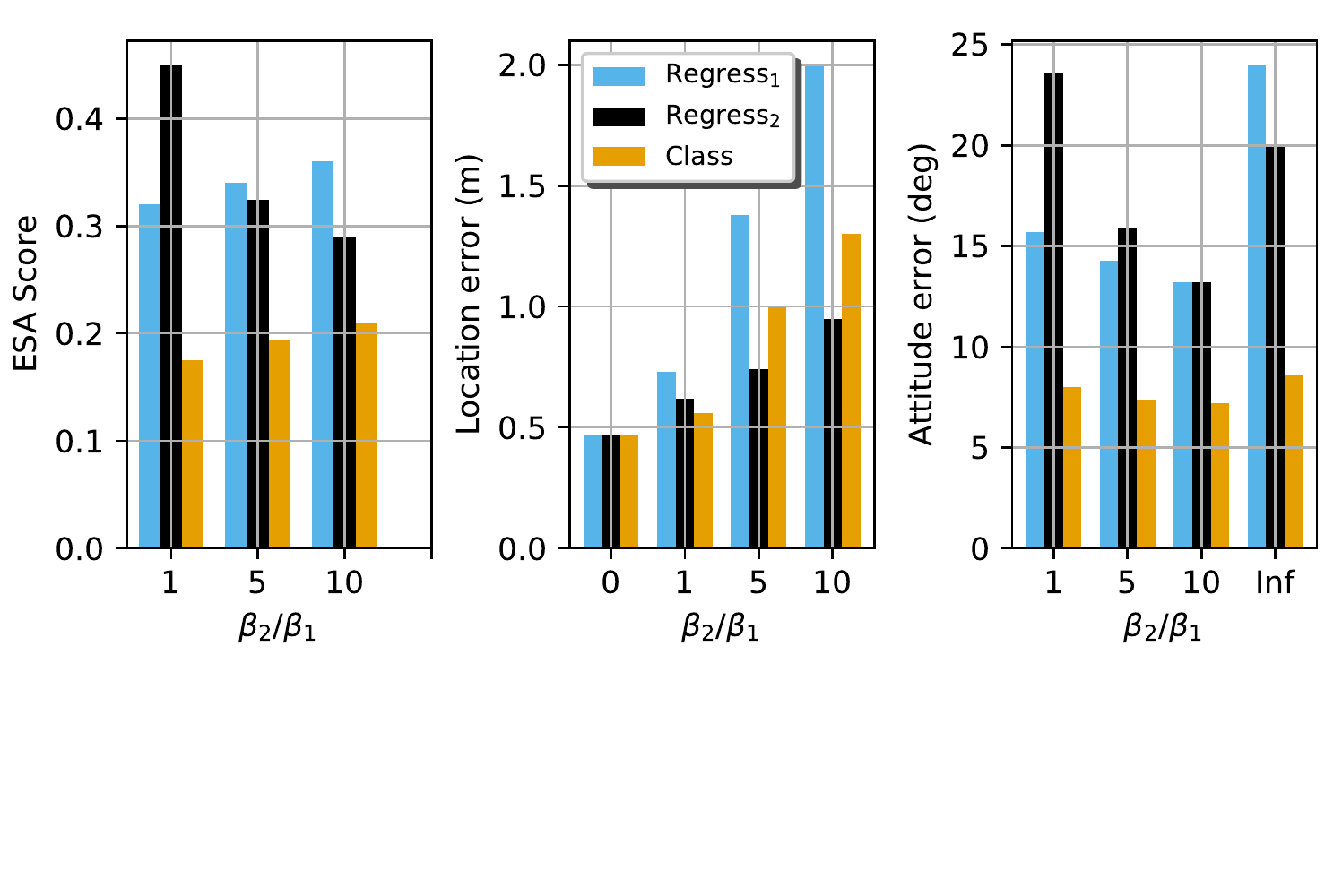}
	\vspace{-40pt}
	\caption{Test errors vs ratio of loss weights. Regress$_1$ and Regress$_2$ regress orientation respectively using the $L_{\alpha}$ and $L_{\cos{\alpha}}$ from Section \ref{sec:ori_reg}.}
	\label{fig:3}
\end{figure}

\begin{minipage}{.45\linewidth}
	\centering
	\vspace{10pt}
	\scriptsize{
		\begin{tabular}{llll}
			\hline
			& &\multicolumn{2}{l}{  Angular error}\\
			\hline
			$\Delta$ & $\#$Bins & Train & Test \\
			\hline
			3     & 16     &     6.5$^{\circ}$      &     55.1$^{\circ}$      \\
			6     & 16     &   5.3$^{\circ}$        &    8.6$^{\circ}$       \\
			9     & 16     &   8.0$^{\circ}$       &      10.3$^{\circ}$     \\ \hline
			6     & 4      &     11.8$^{\circ}$     &      20.0$^{\circ}$     \\
			6     & 8      &      8.9$^{\circ}$      &     11.9$^{\circ}$      \\
			6     & 24     &       3.1$^{\circ}$     &  7.4$^{\circ}$  \\
			\hline
	\end{tabular}}
	\captionof{table}{Impact of orientation soft classification parameters. $\#$Bins is the number of bins per dimension.}
	\label{tab:ori_tuning}
\end{minipage}
\hfill
\begin{minipage}{.45\linewidth}
	\vspace{15pt}
	\centering
	\scriptsize{
		\begin{tabular}{lll}
			\hline
			&\multicolumn{2}{l}{  Angular error}\\
			\hline
			Method & Train & Test \\
			\hline
			Regress$_1$     &     6.7$^{\circ}$      &     13.5$^{\circ}$      \\
			Regress$_2$    &   6.9$^{\circ}$        &    13.4$^{\circ}$       \\
			Regress$_3$    &     9.0$^{\circ}$     &      20.0$^{\circ}$     \\
			Class                   &   5.3$^{\circ}$       &      8.0$^{\circ}$     \\
			\hline
	\end{tabular}}
	\captionof{table}{Orientation error for each method. Regress$_3$ uses regression of 3D points, whereas Regress$_1$ and Regress$_2$ correspond to the best $\beta$ ratio in Fig. \ref{fig:3}.}
	\vspace{10pt}
	\label{tab:ori_overfit}
\end{minipage}

 As one can see, $\Delta$ which is used to scale the Gaussian tail, acts as regularizer: when it is too small, it leads to overfitting, whereas when it is too high, precision is decreased, leading to underfitting. Increasing the number of bins per dimension of the orientation discrete space, improves the precision but the number of network parameters has cubic growth. Furthermore, similarly to $\Delta$, it can lead to overfitting, since bins will be less often activated during training.

Fig. \ref{fig:3} evaluates this method against regressing orientation on \textit{soyuz\_easy}, for different ratios of loss weights. Interestingly, for the three alternatives, using the network only for orientation estimation by setting $\beta_1=0$ in (\ref{eq:loc_total}) yields higher orientation error than performing both tasks simultaneously. The same cannot be said about the location error which grows with $\beta_2$.  Table \ref{tab:ori_overfit} compares the  orientation errors of train and test sets between these methods plus regressing instead three 3D keypoints. We can see that all three regression alternatives are outperformed and suffer from more overfitting on this dataset than the classification approach. It is worth noting that we have experimented using the adaptive weighting based on Laplace likelihood in \cite{kendall2017geometric} but achieved poor results. Moreover, optimal loss weights are subjective to the importance assigned to the specific tasks. \par

\begin{figure}[th]
	\centering
	\begin{tabular}{@{}c@{ }c@{}}
		\includegraphics[scale=0.15]{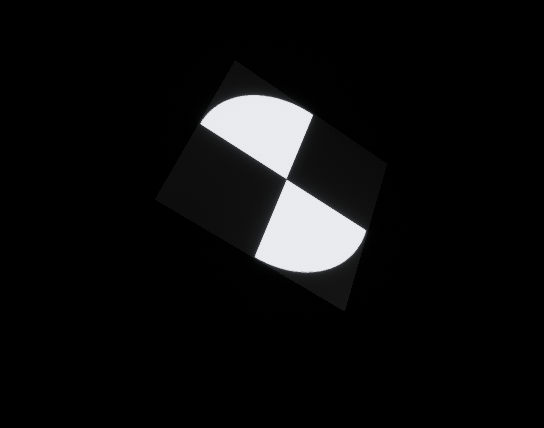} & 
		\includegraphics[scale=0.45]{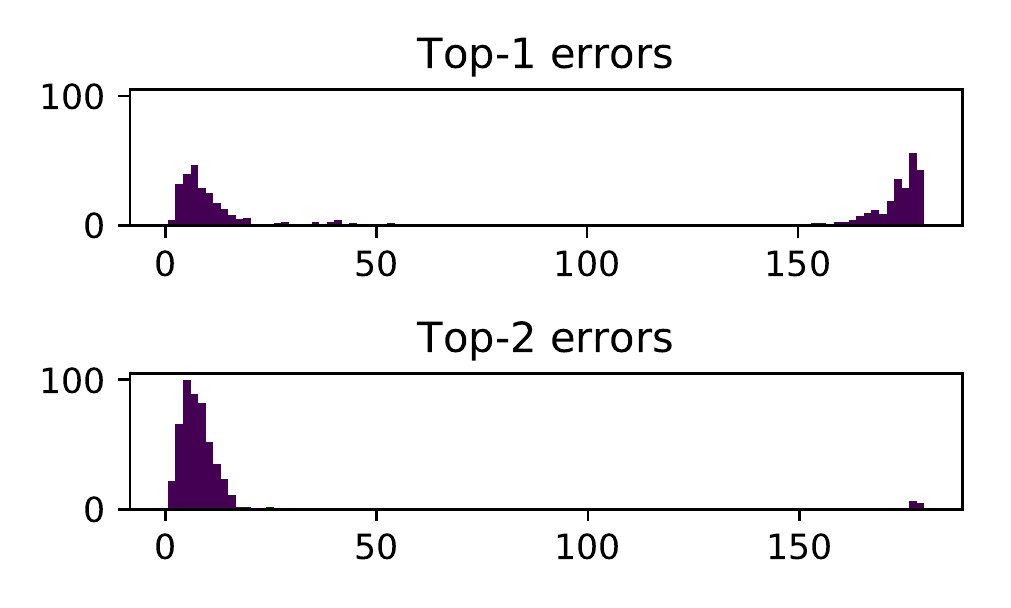}
		\vspace{-5pt}
	\end{tabular}
	\begin{tabular}{@{}c@{}}
		\includegraphics[scale=0.47]{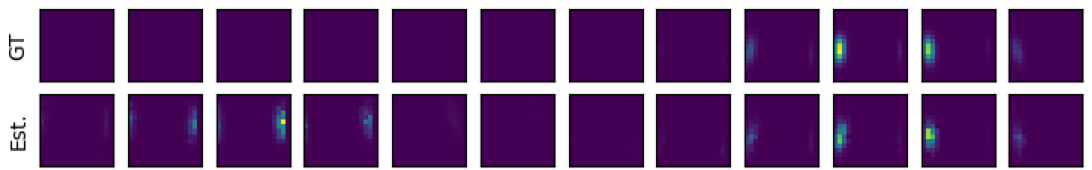}
	\end{tabular}
	\caption{Multimodal orientation estimation experiment with a symmetrical marker, shown on the \textit{top-left}. Histograms of angular errors (deg) are shown on the \textit{top-right} for the testing set: Top-1 error corresponds to our single-hypothesis estimation method, whereas Top-2 error is scored as the hypothesis with smallest error from the top 2 hypothesis estimated by our EM framework. The \textit{bottom} image shows on the top row the encoded label of the left frame, whereas the bottom row shows the respective network output after training.}
	\label{fig:multimodal_ori}
\end{figure}

To demonstrate multimodal orientation estimation, we collected, via URSO, a dataset for the symmetrical marker shown in Fig. \ref{fig:multimodal_ori}. As shown in this figure, after training, the network learns to output two modes representing the two possible solutions. Using naively our unimodal estimation method on this dataset results in the error distribution labeled: \textit{Top-1 errors} in Fig. \ref{fig:multimodal_ori}, whereas if we use the multimodal EM algorithm, proposed in Section \ref{multimodal_EM}, and score the best of two hypothesis: \textit{Top-2} errors, we see that this method finds frequently the right solution. 

\begin{figure}[h]
	\begin{tabular}{@{}c@{ }c@{}}
		\includegraphics[scale=0.47]{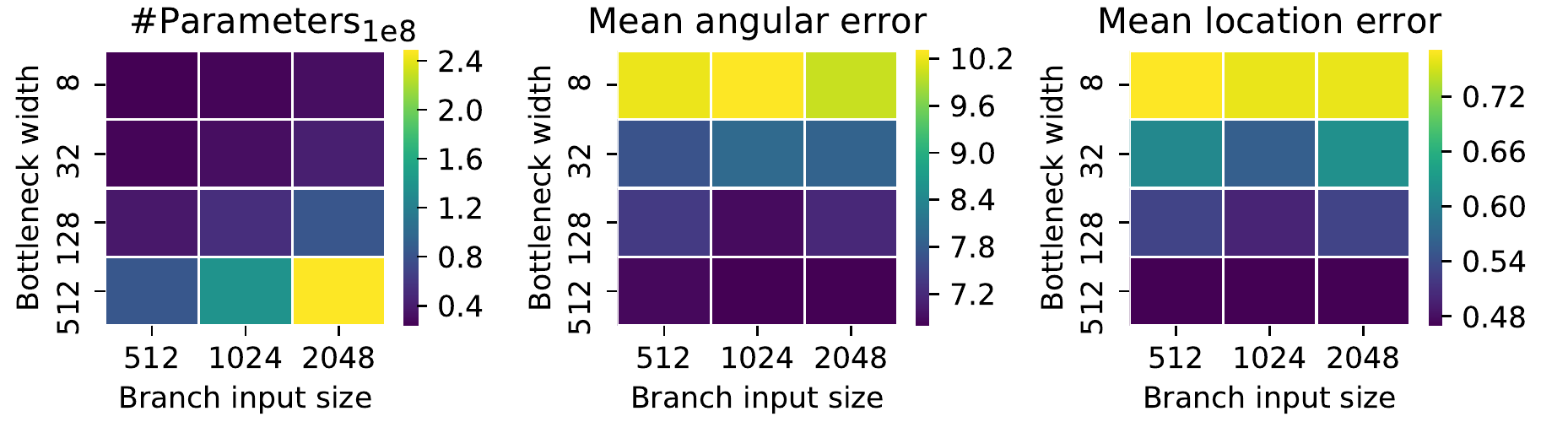}
	\end{tabular}
	\caption{Bottleneck width and size of branch input layers \textit{vs.} performance and complexity in terms of number of parameters on \textit{soyuz\_easy}.}
	\label{fig:bottleneck}
\end{figure}

Fig. \ref{fig:bottleneck} shows how feature compression in the bottleneck layer degrades performance and controls the network size. Similarly, for both tasks, performance changes significantly from using 8 to 128 convolutional filters.

\begin{minipage}{.4\linewidth}
	\vspace{7pt}
	\centering
	\scriptsize{
		\begin{tabular}{llll}
			\hline
			Network & Loc. err. & Ori. err \\
			\hline
			ResNet-18    & 1.7 m         &    19.9$^{\circ}$    \\
			ResNet-34    & 1.4 m        &    20.0$^{\circ}$    \\
			ResNet-50    & 1.1 m        &    13.0$^{\circ}$    \\
			ResNet-101    & 1.0 m       &    12.2$^{\circ}$   \\
			\hline
	\end{tabular}}
	\vspace{10pt}
	\captionof{table}{Impact of architecture depth on \textit{soyuz\_hard}.}
	\vspace{10pt}
	\label{tab:depth}
\end{minipage}%
\hfill
\begin{minipage}{.46\linewidth}
	\vspace{5pt}
	\centering
	\scriptsize{
		\begin{tabular}{lll}
			\hline
			Resolution  & Loc. err. & Ori. err\\
			\hline
			320$\times$240  & 1.6 m & 24.9$^{\circ}$ \\
			640$\times$480  & 1.1 m &  13.0$^{\circ}$      \\
			1280$\times$960 & 1.3 m & 10.7$^{\circ}$     \\
			\hline
	\end{tabular}}
	\vspace{10pt}
	\captionof{table}{Impact of image resolution on \textit{soyuz\_hard}.}
	\label{tab:resolution}
\end{minipage}

\begin{minipage}{.4\linewidth}
	\vspace{-5pt}
	\centering
	\scriptsize{
		\begin{tabular}{llll}
			\hline
			Aug. & Loc err. & Ori err. \\
			\hline
			None    &   1.06 m  &     19.5$^{\circ}$    \\
			Rotation     &   0.56 m	&   8.0$^{\circ}$     \\
			\hline
	\end{tabular}}
	\vspace{10pt}
	\captionof{table}{Impact of applying rotation perturbations on \textit{soyuz easy}}
	\vspace{20pt}
	\label{tab:aug}
\end{minipage}%
\hfill
\begin{minipage}{.5\linewidth}
	\vspace{-5pt}
	\centering
	\scriptsize{
		\begin{tabular}{llll}
			\hline
			Dataset &  Loc err. & Ori err. \\
			\hline
			SPEED &  0.17 m     & 4.0$^{\circ}$	 \\
			Soyuz hard &   0.8 m &     7.7$^{\circ}$ \\
			Dragon hard  &   0.9 m    & 13.9$^{\circ}$\\
			\hline
	\end{tabular}}
	\vspace{10pt}
	\captionof{table}{Results per dataset obtained with 24 bins per orientation dimension and 128 bottleneck filters.}
	\vspace{10pt}
	\label{tab:final}
\end{minipage}
Beyond 128 features, performance gain incurs a great memory footprint. Performance does not seem to be much sensitive to the size of the head input layers. \par
The impact of the architecture depth is shown in Table ~\ref{tab:depth}. ResNet with 50 layers is significantly better than its shallower counterparts, however adding more layers does not seem to improve much more the performance. Table \ref{tab:resolution} shows that orientation estimation is quite sensitive to the image input resolution. The same is not clear for localization.

\begin{figure}[b]
	\centering
	\vspace{-10pt}
	\begin{tabular}{@{}c@{ }c@{}}
		\includegraphics[scale=0.55]{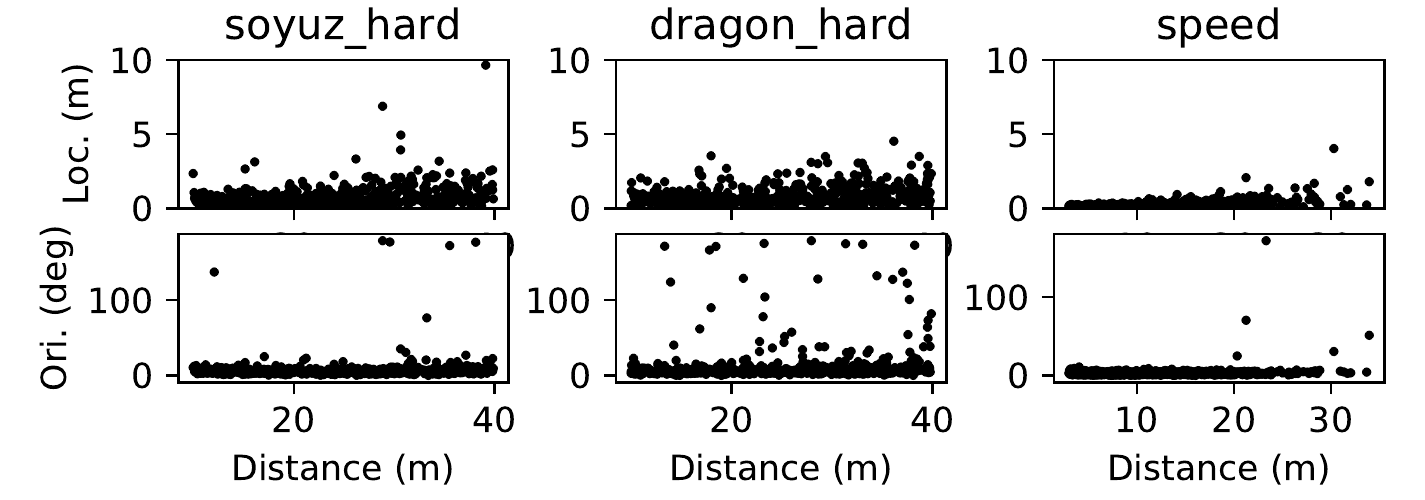}
	\end{tabular}
	\caption{Test-set errors distributed by object distance, for the models reported in Table \ref{tab:final}.}
	\label{fig:err_dists}
\end{figure} 

\begin{figure}[tb]
	\centering
	\begin{tabular}{@{}c@{\quad}c@{}}
		\includegraphics[scale=0.32]{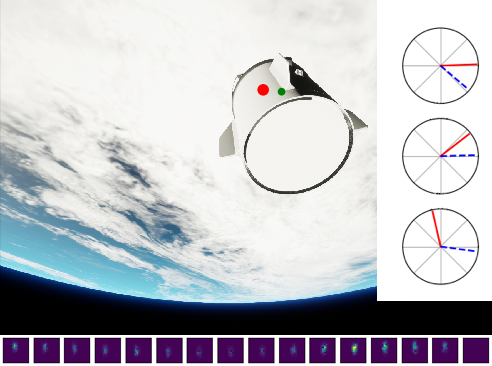}&
		\includegraphics[scale=0.32]{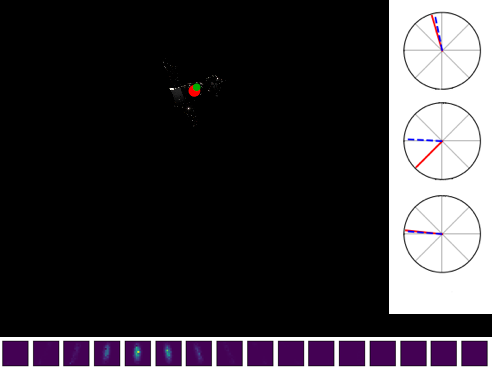}\\[-5pt]
		\scriptsize{(a)} & \scriptsize{(b)} \\
		\includegraphics[scale=0.32]{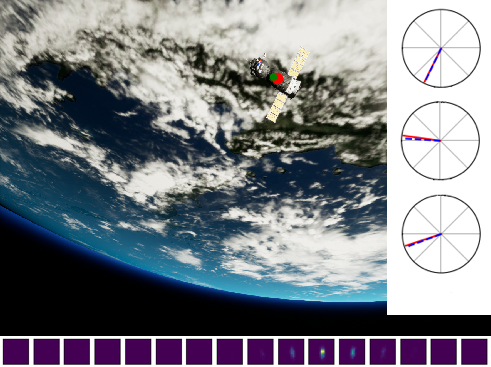}&
		\includegraphics[scale=0.32]{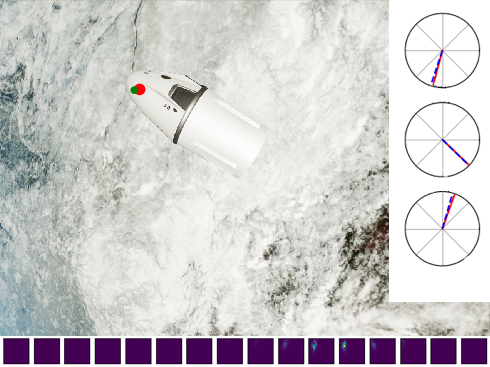} \\[-5pt]
		\scriptsize{(c)} & \scriptsize{(d)} \\
		\includegraphics[scale=0.32]{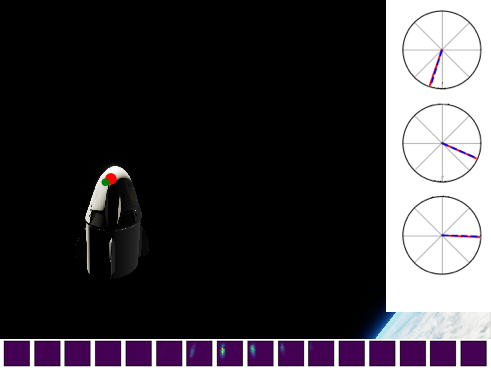}&
		\includegraphics[scale=0.32]{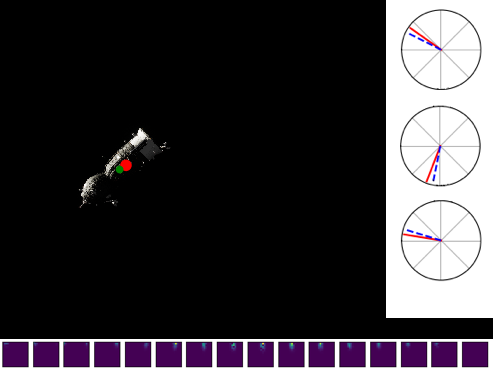}\\[-5pt]
		\scriptsize{(e)} & \scriptsize{(f)}  \\[-5pt]
		
	\end{tabular}
	\caption{Failure and success cases from our testing sets with predicted and groundtruth poses, and orientation weights. Predicted and labeled 2D position are shown respectively as green and red dots. Predicted and labeled orientations are shown in the polar plots as Euler angles. (a) Incorrect orientation due to an ambiguous view. Notice how the respective distribution of weights is more spread out. (b) Poor orientation estimation due to poor lighting. (c) and (d) Good results under challenging background. }
	\label{fig:final_examples}
\end{figure}

In terms of data augmentation, as reported in Table \ref{tab:aug}, rotation perturbations prove to be an effective technique to augment the dataset and our sim-to-real augmentation is essential to apply models learned on URSO to real footage as shown in {\scriptsize \url{https://youtu.be/x8IbxmOz730}}, particularly to deal with the lighting changes in Fig. \ref{fig:aug}. Furthermore, as shown in Table \ref{tab:final_esa_results}, we achieved \nth{2} place on the real dataset just by using our sim-to-real augmentation pipeline with the 5 real images provided. Table \ref{tab:final} compares performance between the three datasets using an increased bottleneck width and orientation output resolution. As we can see, SPEED with better lighting conditions is the easiest dataset and \textit{dragon\_hard} is the most challenging dataset due to viewpoint ambiguity, as shown in  Fig. \ref{fig:err_dists} and Fig. \ref{fig:final_examples}.a.

\begin{table}[t]
	\vspace{10pt}
	\centering
	\scriptsize{
		\begin{tabular}{lll}
			\hline
			Team & Real err. & Synthetic err. \\
			\hline
			UniAdelaide &     0.3752      &     0.0095      \\
			EPFL\_cvlab    &   0.1140      &      0.0215    \\ \hline
			Triple ensemble (ours)    &     0.1555     &      0.0571     \\
			Best model $\dagger$ (ours)    &      0.1630     &     0.0604     \\
			\hline
			Top 10 average &      1.3848     &     0.1515 \\
			\hline
	\end{tabular}}
	\caption{ESA pose estimation final scores of top 3 teams. Results for $\dagger$  were obtained for 20 $\%$ of the full test set. For the complete leaderboard, refer to \cite{ESAChallenge}.}
	\label{tab:final_esa_results}
\end{table}

Table \ref{tab:final_esa_results} summarizes the results of the ESA pose estimation challenge. Our best single model used a bottleneck width of 800 filters and 64 bins per orientation dimension and was trained for a total of 500 epochs, whereas our second best model using 512 bottleneck filters and 32$\times$32$\times$32 orientation bins achieved respectively: 0.144 and 0.067 on the real and synthetic set. To combine the higher precision of the best model with the less likely overfitting second model we used a triple ensemble, which
is an average of results (using quaternion averaging) of this last model plus two models with 64$\times$64$\times$64 bins, picked at different training epochs. Our accuracy comes with a very large amount of parameters (around 500M) and it is still far from the scores of the top 2 teams, which rely on 2D keypoint regression solutions, image cropping+zooming and robust P$n$P. As shown in Fig. \ref{fig:err_dists}, gross errors start appearing after 20 m, therefore we could also benefit from running the models a second time on zoomed images, since we only used half the original size. 

\subsection{Conclusion and Future Work}

This paper proposed both a simulator and a DL framework for spacecraft pose estimation. Experiments with this framework reveal the impact of several network hyperparameters and training choices and attempts to answer open questions, such as, what is the best way to estimate orientation?
We conclude that estimating orientation based on soft classification gives better results than direct regression and furthermore it provides the means to model uncertainty. This information is useful not only to make decisions but it can be used for filtering the pose if a temporal sequence is provided.
A promising direction is to address tracking using Recurrent Neural Networks and video sequences generated using URSO. As future work, we also plan to extend URSO to SLAM to address targets with unknown geometry. 
\par
The architecture proposed in this work is not scalable in terms of image and orientation resolution. Future work should consider how to replace the dense connections without sacrificing performance, e.g., pruning the last layer connections. Additionally, the results reported in this work were obtained using a dedicated network for each dataset. It may be beneficial sharing the same backbone in terms of efficiency and performance.

\subsection{Acknowledgments}

This work is supported by grant EP/R026092 (FAIR-SPACE Hub) through UKRI under the Industry Strategic Challenge Fund (ISCF) for Robotics and AI Hubs in Extreme and Hazardous Environments. The authors are also grateful for the feedback and discussions with
Peter Blacker, Angadh Nanjangud and Zhou Hao.

\bibliographystyle{ieeetr} 
{\footnotesize
\bibliography{root}
}

\end{document}